\documentclass{article}

\usepackage{microtype}
\usepackage{graphicx}
\usepackage{subcaption}
\usepackage{booktabs} 

\usepackage{graphicx}
\usepackage{booktabs}
\usepackage{multirow}
\usepackage[table]{xcolor}
\usepackage{arydshln}
\usepackage{tcolorbox}
\usepackage{hyperref}

\usepackage[preprint]{icml2026}

\usepackage{amsmath}
\usepackage{amssymb}
\usepackage{mathtools}
\usepackage{amsthm}

\newcommand{\Sspace}{\mathcal{S}}
\newcommand{\Aspace}{\mathcal{A}}
\newcommand{\Tspace}{\mathcal{T}}
\newcommand{\Rspace}{\mathcal{R}}
\newcommand{\Ospace}{\Omega}
\newcommand{\Ofunc}{\mathcal{O}}

\usepackage[capitalize,noabbrev]{cleveref}

\theoremstyle{plain}

\theoremstyle{definition}

\theoremstyle{remark}

\newcommand{\remove}[1]{}

\newcommand{\DataAgentName}{AReaL-SEA}

\usepackage[textsize=tiny]{todonotes}

\usepackage{tcolorbox}
\tcbuselibrary{breakable,skins}
\tcbuselibrary{listings}
\usepackage{listings}

\newtcolorbox{llmexample}[2][]{
  enhanced,
  breakable,
  colback=gray!3,
  colframe=gray!50,
  fonttitle=\bfseries,
  title={#2},
  boxrule=0.5pt,
  arc=2pt,
  left=6pt,
  right=6pt,
  top=6pt,
  bottom=6pt,
  #1
}

\lstdefinestyle{jsonstyle}{
  basicstyle=\ttfamily\small,
  breaklines=true,
  columns=fullflexible,
  keepspaces=true,
  showstringspaces=false,
  frame=single,
  framerule=0.3pt
}

\lstdefinestyle{markdownstyle}{
  basicstyle=\ttfamily\small,
  breaklines=true,
  columns=fullflexible,
  keepspaces=true,
  showstringspaces=false,
}

\lstdefinestyle{xmlstyle}{
  basicstyle=\ttfamily\small,
  breaklines=true,
  columns=fullflexible,
  keepspaces=true,
  showstringspaces=false,
}

\lstdefinestyle{txtstyle}{
  basicstyle=\ttfamily\small,
  breaklines=true,
  columns=fullflexible,
  keepspaces=true,
  showstringspaces=false,
}

\lstdefinestyle{plainstyle}{
  basicstyle=\ttfamily\small,
  breaklines=true,
  columns=fullflexible,
  keepspaces=true,
  showstringspaces=false,
}

\lstdefinestyle{plaintextstyle}{
  basicstyle=\ttfamily\small,
  breaklines=true,
  columns=fullflexible,
  keepspaces=true,
  showstringspaces=false,
}

\icmltitlerunning{From Self-Evolving Data to Verifiable-Reward RL for Interactive Tool-Using Agents}

\begin{document}

\twocolumn[
  \icmltitle{From Self-Evolving Synthetic Data to Verifiable-Reward RL: Post-Training Multi-turn Interactive Tool-Using Agents}



  \icmlsetsymbol{equal}{*}
  
\begin{icmlauthorlist}
    \icmlauthor{Jiaxuan Gao}{thu,equal}
    \icmlauthor{Jiaao Chen}{eigen,equal}
    \icmlauthor{Chuyi He}{ind}
    \icmlauthor{Shusheng Xu}{thu}
    \icmlauthor{Di Jin}{eigen}
    \icmlauthor{Yi Wu}{thu}
\end{icmlauthorlist}

\icmlaffiliation{thu}{Tsinghua University}
\icmlaffiliation{eigen}{Eigen AI}
\icmlaffiliation{ind}{Independent Researcher}

  \icmlcorrespondingauthor{Di Jin}{di@eigenai.com}
  \icmlcorrespondingauthor{Yi Wu}{jxwuyi@gmail.com}


  \vskip 0.3in
]



\printAffiliationsAndNotice{\icmlEqualContribution}

\begin{abstract}
Interactive tool-using agents must solve real-world tasks via multi-turn interaction with both humans and external environments, requiring dialogue state tracking, multi-step tool execution, while following complex instructions. Post-training such agents is challenging because synthesis for high-quality multi-turn tool-use data is difficult to scale, and reinforcement learning (RL) could face noisy signals caused by user simulation, leading to degraded training efficiency.
We propose a unified framework that combines a \emph{self-evolving} data agent with verifier-based RL. Our system, \textbf{\DataAgentName{}}, is a hierarchical multi-agent engine that synthesizes tool-grounded dialogues together with \emph{executable} per-instance checkers, and improves generation reliability via closed-loop self-evolving process that updates prompts and workflow. Building on the synthetic data, we develop an RL recipe that first fine-tunes the user model and then applies GRPO-style training with trajectory-level group-relative advantages and dynamic filtering, yielding consistent improvements beyond SFT. Evaluated on \(\tau^{2}\)-bench, our best model reaches \(73.0\%\) pass\^{}1 on Airline and \(98.3\%\) pass\^{}1 on Telecom, matching or exceeding frontier models. Overall, our results suggest a scalable pathway for bootstrapping complex tool-using behaviors without expensive human annotation.\footnote{Code and data are open-sourced in \href{https://github.com/inclusionAI/AReaL/tree/main/examples/tau2}{https://github.com/inclusionAI/AReaL/tree/main/examples/tau2}.}
\end{abstract}
\section{Introduction}

Large language models (LLMs) have rapidly evolved from generic next-token predictors into broadly capable systems that can follow instructions, reason over long contexts, and support downstream adaptation via fine-tuning and post-training, including strong open-weight model families \citep{tang2025eigen,liu2025deepseek,kimiteam2025kimik2,zhu2026toward}. 
As these models have been integrated into real applications, there has been a central shift from \emph{static question answering} to \emph{interactive task completion}, where the model must 
communicate with humans and interact with the external environment through tool/API calls
to accomplish complex tasks
~\citep{anthropic_claude_opus_45_system_card_2025,yao2024tau,barres2025tau2benchevaluatingconversationalagents, froger2025arescalingagentenvironments}. 

Prior work on tool-augmented agents has largely focused on settings where the agent invokes tools to fulfill self-contained user queries \citep{yehudai2025surveyevaluationllmbasedagents}. In contrast, interactive agents operate in the presence of an \emph{active user} throughout the interaction—a distinction that introduces two fundamental challenges.
First, \textbf{critical information resides on the user side}: unlike single-turn tool-use where all necessary context is provided upfront, interactive agents must actively elicit user preferences and private details through multi-turn dialogue before taking actions.
Second, \textbf{user behavior is inherently uncertain}: users may provide information incrementally, change their minds, or respond in unexpected ways.
For instance, in the \(\tau\)-bench airline~\citep{yao2024tau}, an agent handling a flight change request must (i) ask clarifying questions for the user's preferences, (ii) query database via API calls to find suitable alternatives, (iii) verify that the proposed change complies with policies, and (iv) execute the modification.


Despite the availability of capable open-weight foundation models, post-training these models into effective interactive agents remains challenging~\citep{barres2025tau2benchevaluatingconversationalagents}. We identify two primary bottlenecks. 
The first bottleneck is \textbf{scalable data acquisition}. High-quality training data for multi-turn tool-using dialogues is difficult to obtain at scale. Human annotation requires substantial effort, particularly for scenarios involving complex domain constraints. Automated synthesis is also challenging: generating  \emph{sufficiently challenging} tasks for effective post-training requires satisfying intricate domain rules while simultaneously providing a simulated user with coherent instructions and private information ~\citep{barres2025tau2benchevaluatingconversationalagents,prabhakar2025apigenmtagenticpipelinemultiturn,xu2025toucan}.
The second bottleneck is \textbf{reinforcement learning for interactive agents}. 
Since interactive tasks require a user to drive the conversation, RL training must incorporate a user simulator, introducing additional non-deterministic dynamics into the rollout process~\citep{barres2025tau2benchevaluatingconversationalagents}.
Furthermore, in cases \emph{users can also invoke tool calls} as in the dual-control setting of $\tau^{2}$-bench~\citep{barres2025tau2benchevaluatingconversationalagents}, we find that open-source models exhibit unstable behavior when simulating such tool-using users. This instability in user behaviors significantly undermine the success rate of rollouts and introduces noisy training signals.

We address these challenges with a two-part post-training framework.
First, we present \textbf{\DataAgentName{}}, a hierarchical \emph{self-evolving} multi-agent system that autonomously generates and validates training data with minimal human supervision.
\DataAgentName{} comprises an \emph{orchestration layer} that designs workflows, writes agent prompts, and drives iterative self-evolution, as well as an \emph{execution layer} of specialized worker agents that synthesize tasks, interaction trajectories, and \emph{executable} per-instance verification functions that server as reward signals for RL.
Second, we develop an \textbf{Reinforcement Learning recipe} for multi-turn interactive tool use where user behavior introduces substantial variance. A key prerequisite is preparing a reliable user simulator. We find that off-the-shelf models exhibit unstable behavior when simulating tool-using users. We therefore first fine-tune the user model via SFT to ensure stable, instruction-following behavior before using it in RL rollouts. For training, we employ GRPO~\citep{shao2024deepseekmathpushinglimitsmathematical} with large batch sizes and dynamic sampling to stabilize learning under the inherent uncertainty of user-driven interactions. We use verifier-based outcome rewards based on the generated verification functions by EiganData: the resulting state is compared against the ground-truth final state to determine success.


Our framework produces substantial improvements across all three domains of \(\tau^{2}\)-bench using Qwen3 backbones.
For Qwen3-30B-A3B-2507, SFT alone yields strong gains, for example, improving Telecom from $p^{1}=28.5$ to $85.4$. RL training further boosts performance, achieving $p^{1}=70.5$ on Airline, $75.0$ on Retail, and $95.6$ on Telecom.
For Qwen3-235B-A22B-2507, RL reaches \(p^{1}=73.0\) on Airline, \(75.0\) on Retail, and \(98.3\) on Telecom, matching or exceeding frontier model in all domains.
Taken together, our findings demonstrate that self-evolving synthetic data generation, combined with stabilized verifier-based RL, can reliably improve multi-turn tool-use capabilities.

\noindent\textbf{Our key contributions are as follows:}
\begin{itemize}
\item \textbf{\DataAgentName{}}, a \emph{self-evolving} data synthesis system that generates verifiable, complex and high-quality multi-turn tool-use training instances.
\item \textbf{An RL recipe for interactive tool-use agents}, comprising user model fine-tuning, large-batch training to mitigate user behavior variance, dynamic sampling, and verifier-based outcome rewards.
\item \textbf{Extensive empirical evaluation}, including ablations quantifying the contributions of each component. Our approach achieves state-of-the-art results on \(\tau^{2}\)-bench using fully open-weight models.
\end{itemize}

\section{Related Work}

\textbf{Tool-using Language Agents.}
Tool-using agents have emerged as a paradigm for extending LLMs beyond their parametric knowledge with external tools such as APIs, web browsers, and search engines~\cite{schick2023toolformer,yao2022react,parisi2022talmtoolaugmentedlanguage,liang2023taskmatrixaicompletingtasksconnecting,paranjape2023artautomaticmultistepreasoning}.
ReAct~\cite{yao2022react} introduced interleaved reasoning traces and actions to ground model outputs in external knowledge.
Toolformer~\cite{schick2023toolformer} demonstrated that LLMs can teach themselves when and how to invoke tools through self-supervised learning.
Search agents~\cite{nakano2022webgptbrowserassistedquestionansweringhuman,gur2024realworldwebagentplanninglong,gao2025beyond,jin2025search} extend these capabilities to autonomous web navigation question answering.
To evaluate tool-use capabilities at scale, ToolLLM~\cite{qin2023toolllm}, BFCL~\cite{patilberkeley} and ACEBench~\cite{chen2025acebench} provide realistic evaluation of function calling across serial, parallel, and multi-turn interactions.
While these benchmarks measure isolated tool execution, $\tau$ and $\tau^2$-bench~\cite{yao2024tau,barres2025tau2benchevaluatingconversationalagents} advanced evaluation toward realistic conversational scenarios by testing interaction between simulated users and agents.



\textbf{Synthetic Data Generation.}
Synthesis data plays a key role in providing scalable training data with minimal human efforts ~\citep{wang2023self,yu2025cotselfinstructbuildinghighqualitysynthetic,xu2025wizardlmempoweringlargepretrained,xu2024magpiealignmentdatasynthesis,chen2024dynamicskilladaptationlarge,robeyns2025selfimprovingcodingagent}. Recent work moves toward \emph{trajectory synthesis} with stronger grounding and validation. 
APIGen~\cite{liu2024apigenautomatedpipelinegenerating} generates function-calling data with execution-based checks, and APIGen-MT~\cite{prabhakar2025apigenmtagenticpipelinemultiturn} extends this to multi-turn tool-use trajectories via simulated agent--human interplay and reviewer-style validation. 
TOUCAN~\cite{xu2025toucan} scales further by synthesizing 1.5M tool-agent trajectories from hundreds of real MCP environments.
Supporting infrastructure for multi-agent generation has also matured (e.g., hierarchical workflow~\citep{liu2025hierarchicalmultiagentworkflowszeroshot} and unified codebases~\cite{ye2025maslabunifiedcomprehensivecodebase}). 
In contrast to prior static pipelines that primarily target SFT data, our data engine dynamically designs workflow and evolves through feedback: it learns from its own failures and produces verifiers that enable RL training from the same synthetic data.


\textbf{Reinforcement Learning for LLM.}
Reinforcement learning has emerged as a central technique for post-training LLMs, demonstrating significant effectiveness in enhancing model capabilities. Reinforcement Learning from Human Feedback (RLHF)~\cite{ouyang2022training} learns a reward model to model human preferences and has successfully improved alignment and instruction-following abilities. 
Beyond RLHF, the paradigm shift toward Reinforcement Learning with Verifiable Rewards (RLVR), 
which uses programmatic verifiers to evaluate the quality of model outputs, significantly enhancing model capability on complex reasoning tasks
~\cite{liu2024deepseek,li2025simulatingenvironmentsreasoningmodels,zhang2026arenarl}. Building upon RLVR, recent advances in agentic RL further applies RL on training long-horizon tool-using agents~\cite{liu2025gemgymagenticllms,lu2025scaling,zhuang2025workforceagent,gao2025beyond,fu2025areal}. 
Our work follows the agentic RL paradigm and investigate RL for interactive tool-using agents. Unlike these previous works, user simulation in the interactive setting brings noises to RL training. We propose user model fine-tuning as a critical component for training interactive tool-using agents.

\section{Preliminary}


\subsection{Interactive Tool-Using Agents} 
We formulate the problem as a decentralized partially observable Markov decision process (Dec-POMDP), defined by the tuple $\mathcal{M} = \langle \Sspace, \Aspace, \Tspace, \Rspace, \Ospace, \Ofunc, \gamma, \rho_0\rangle$ with two players, i.e. the agent and the user. The agent uses a parameterized policy $\pi_\theta$ and the user policy is denoted as $\pi_{user}$.

The state $s_t \in \Sspace$  captures all environment information. 
The initial state distribution $\rho_0$ is defined as $s_0 \sim \rho_0(\cdot|q)$ where $q\in \mathcal D$ is the task specification.
$\Aspace$ is the set of possible actions for each player. This may include provided tool calls, or messages between the agent and the user. 
$\Tspace: \Sspace \times \Aspace \rightarrow \Delta(\Sspace)$ is the transition function, which models the state updates caused by actions such as tool calls.
At each turn $t$, the agent and the user both observe a local observation, $o_t^{agent}=\Ofunc(s_t;agent)\in \Ospace$ and $o_t^{user}=\Ofunc(s_t;user)\in \Ospace$, respectively. 
For tool-calling agents, the observation usually includes system context that entails task instructions, available tool specifications, and the interaction history.

For reward function $\Rspace$, we employ an outcome reward setting where the agent receives feedback only upon task completion. The reward function $\Rspace: \Sspace \times \Aspace \rightarrow \mathbb{R}$ is defined as $\Rspace(s_t, a_t) = R(s_T)$ for $t=T$ and $\Rspace(s_t,a_t)=0$ for $t<T$, where $T$ denotes the terminal turn and $R(s_T)$ is a scalar metric evaluating the correctness of the final state $s_T$. 


\subsection{Policy and Objective}

\textbf{Agent \& User Policy.} The agent policy is an LLM $\pi_\theta$ that outputs a text $y_t$ given observation $o_t^{agent}$. The output $y_t$ contains both the agent's reasoning about the current state and actions to take, e.g. tool calls. Agent action $a^{agent}_t$ is then extracted from $y_t$ through a heuristic parsing function, e.g. by matching the JSON-format content. The LLM  generates tokens in an autoregressive manner, $\pi_\theta(y_t \mid o_t^{agent}) = \prod_{i=1}^{|y_t|} \pi_\theta(y_{t,i} \mid o_t^{agent}, y_{t, <i})$.
Similarly, the user policy $\pi_{user}$ generates actions $a^{user}_t$ conditioned on its observation $o^{user}_t$. In interactive tool-using agent settings, either the agent or the user can act on each turn. non-acting party's action would be set as $\emptyset$ and, in practice, only the acting party would invoke an LLM generation call.

\textbf{Objective.} The agent's goal is to find policy $\theta$ that maximize the expected reward: $\max_\theta \; J(\theta) = \mathbb{E}_{s_0 \sim {\rho_0(\cdot)}, \tau \sim P(\tau \mid s_0, \pi_\theta,\pi_{user})} \left[ R(s_T) \right]$
where $\tau = (s_0, \textbf{a}_0, \ldots, s_T, \textbf{a}_T)$ is the full trajectory, $T$ is the terminal turn and $\textbf{a}_t=(a^{agent}_t,a^{user}_t)$ is the joint action at turn $t$.

\section{\DataAgentName{}: Self-Evolving Data Synthesis Framework}
\label{sec:data_engine}

\begin{figure*}[htbp]
    \centering
    \includegraphics[width=0.85\linewidth]{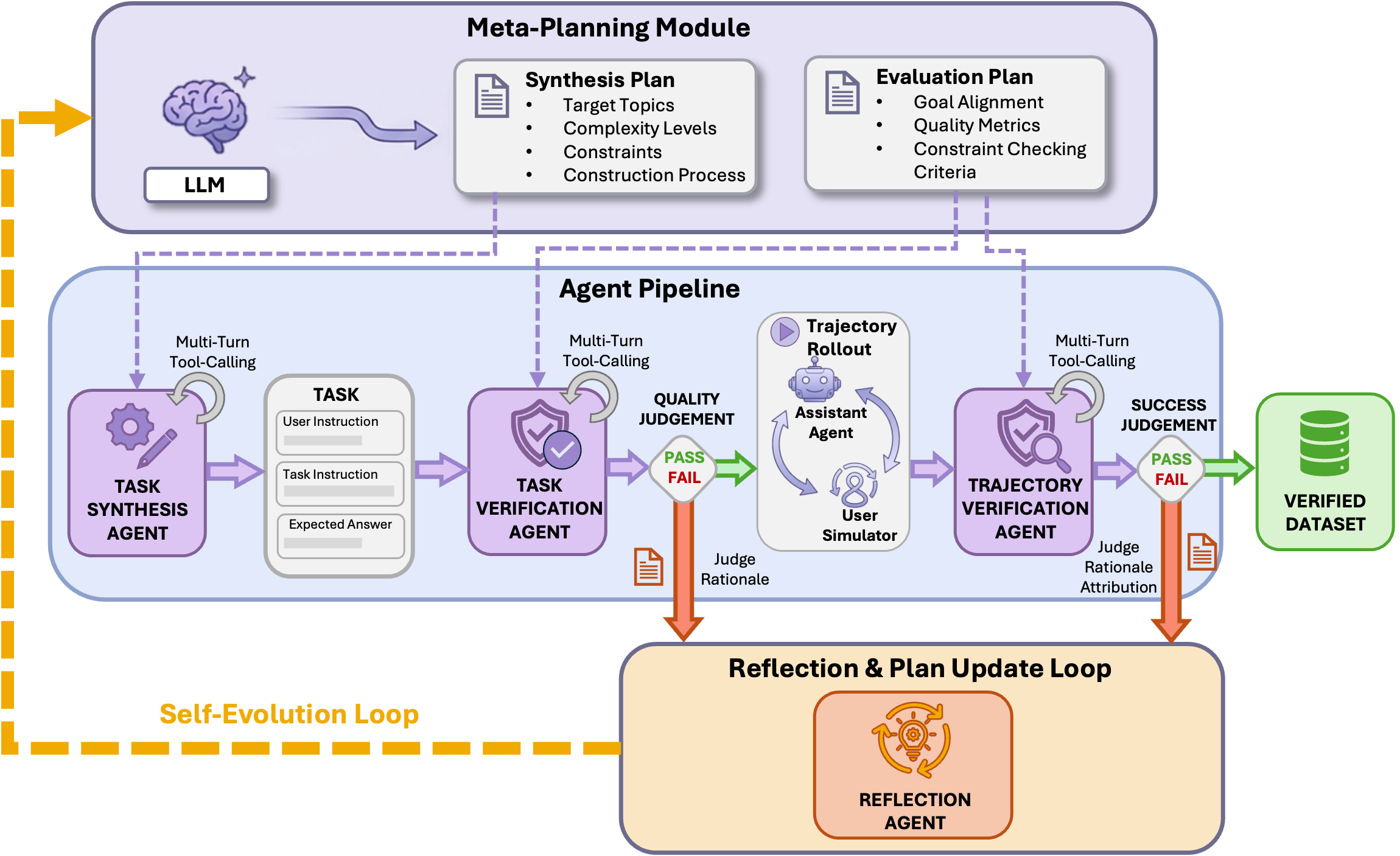}
    \caption{\textbf{Overview of the self-evolving data synthesis framework.} The meta-planning module (\textit{top}) first generates a diversified set of synthesis--evaluation plan pairs, each of which independently enters the agent pipeline (\textit{center}). The pipeline consists of four sequential stages: a \textit{Task Synthesis Agent} produces structured candidate tasks, a \textit{Task Verification Agent} filters them against the evaluation plan, a \textit{Trajectory Rollout Module} simulates multi-turn assistant--user interactions on verified tasks, and a \textit{Trajectory Verification Agent} assesses trajectory quality with root-cause attribution. Successfully verified trajectories (green path) are collected into the final dataset, while failures (red paths) are routed to the \textit{Reflection Module} (\textit{bottom}), which analyzes failure patterns and updates both the synthesis and evaluation plans (dashed arrows), closing the self-evolution loop.}
    \label{fig:data_engine}
    \vspace{-5mm}
\end{figure*}

We introduce a self-evolving data synthesis framework that automatically generates, verifies, and refines high-quality training tasks and trajectories through multi-agent orchestration\footnote{\DataAgentName{} is built upon an early version of EigenData~\citep{chen2026eigendata}.}. As illustrated in Figure~\ref{fig:data_engine}, our pipeline consists of four interconnected components: a meta-planning module, a sequential agent pipeline for synthesis and verification, trajectory collection, and a reflection loop that enables iterative self-improvement of the generation process. To achieve diversity, we first generate a broad collection of distinct synthesis plans spanning different domains and task types, then execute each plan through the self-evolving pipeline independently.

\subsection{Diversified Plan Generation}
\label{subsec:diversity}

Generating a diverse training set requires coverage across domains, tool-use patterns, complexity levels, and interaction styles. Rather than relying on a single monolithic plan and hoping for emergent diversity, we explicitly construct diversity at the planning stage.

Concretely, we prompt the meta-planning LLM to sequentially generate $N$ distinct synthesis--evaluation plan pairs $\{(\mathcal{P}_s^{(n,0)}, \mathcal{P}_e^{(n,0)})\}_{n=1}^{N}$. Each synthesis plan $\mathcal{P}_s^{(n,0)}$ specifies a unique combination of target domain (e.g., data analysis, web navigation, software engineering), task complexity tier, required tool-use patterns, and stylistic constraints on user instructions. The corresponding evaluation plan $\mathcal{P}_e^{(n,0)}$ defines domain-appropriate quality rubrics and failure taxonomies tailored to the characteristics of that plan. During sequential generation, previously generated plans are provided as context to the LLM to encourage complementary coverage and minimize redundancy across the plan set.

Once all $N$ plan pairs are instantiated, each enters the self-evolving pipeline (Sections~\ref{subsec:agent_pipeline}--\ref{subsec:reflection}) as an independent stream. This design offers two key advantages. First, it decouples diversity from the stochasticity of any single generation process, providing explicit and controllable coverage guarantees. Second, it allows the reflection loop to specialize its refinements per plan. The final dataset is the union of verified trajectories across all streams: $\mathcal{D} = \bigcup_{n=1}^{N} \mathcal{D}^{(n)}$.

\subsection{Meta-Planning}
\label{subsec:meta_planning}

Each plan pair $(\mathcal{P}_s^{(n,0)}, \mathcal{P}_e^{(n,0)})$ produced in Section~\ref{subsec:diversity} serves as the initial configuration for one stream of the pipeline. The synthesis plan $\mathcal{P}^{(n,0)}_s$ specifies the distribution of tasks to generate, including target domains, complexity levels, required tool-use patterns, and diversity constraints. The evaluation plan $\mathcal{P}^{(n,0)}_e$ defines quality criteria, rubrics, and failure taxonomies used by downstream verification agents. Both plans are initialized by the meta-planning LLM and subsequently refined through the reflection loop described in Section~\ref{subsec:reflection}.

\subsection{Agent Pipeline}
\label{subsec:agent_pipeline}

Given a plan pair $(\mathcal{P}^{(n,k)}_s, \mathcal{P}^{(n,k)}_e)$, our synthesis pipeline processes data through four sequential stages:

\paragraph{Task Synthesis.} A task synthesis agent, guided by $\mathcal{P}^{(n,k)}_s$, generates candidate tasks through a multi-turn tool-calling loop. Each task $q$ is a structured tuple $q = (u, t, a^*)$, where $u$ denotes the user instruction, $t$ the detailed task specification, and $a^*$ the expected answer or completion criteria. The agent iteratively refines its outputs using tool calls to ensure the generated tasks satisfy the constraints specified in the synthesis plan.

\paragraph{Task Verification.} A dedicated verification agent evaluates each candidate task against the evaluation plan $\mathcal{P}^{(n,k)}_e$. Operating through its own multi-turn tool-calling loop, it produces a quality judgment $j_t \in \{\textsc{Pass}, \textsc{Fail}\}$ accompanied by a detailed rationale. Tasks that pass proceed to trajectory rollout; failed tasks, along with their failure rationales, are routed to the reflection module.

\paragraph{Trajectory Rollout.} Verified tasks are executed through a simulated multi-turn interaction between two sub-agents: an \textit{assistant agent} that attempts to solve the task, and a \textit{user simulator} that provides realistic follow-up instructions, clarifications, and feedback. Their interaction produces a complete trajectory $\tau$. This design ensures the resulting trajectories capture the complexity of realistic multi-turn interactions rather than single-turn completions.

\paragraph{Trajectory Verification.} A trajectory verification agent evaluates the complete trajectory $\tau$ against $\mathcal{P}^{(n,k)}_e$, producing a success judgment $j_\tau \in \{\textsc{Success}, \textsc{Fail}\}$ with rationale. Critically, upon failure, this agent also produces an \textit{attribution tag} $c \in \{\textsc{Task}, \textsc{Trajectory}\}$ that identifies whether the root cause lies in a poorly specified task or in the assistant's execution. Successfully verified trajectories are added to the dataset $\mathcal{D}$, while failures are routed to the reflection module with their attribution.

\subsection{Reflection and Self-Evolution}
\label{subsec:reflection}

The reflection module is the key mechanism enabling our framework to \textit{self-evolve}. It aggregates failure cases from both verification stages, each accompanied by detailed rationales and, in the case of trajectory failures, attribution tags. A reflection agent analyzes these failure patterns and generates targeted revisions to both the synthesis plan and the evaluation plan:
\begin{align*}
    &\mathcal{P}_s^{(n,k+1)}, \mathcal{P}_e^{(n,k+1)} = \\&\textsc{Reflect}\!\left(\mathcal{P}_s^{(n,k)}, \mathcal{P}_e^{(n,k)}, \{(q_i, r_i, c_i)\}_{i \in \mathcal{F}^{(k)}}\right),
\end{align*}
where $\mathcal{F}^{(k)}$ is the set of failures at iteration $k$, and $r_i$, $c_i$ denote the rationale and attribution for each failure. This creates a closed loop: systematic failures caused by underspecified task descriptions lead to more precise synthesis plans, while failures caused by overly lenient or strict evaluation criteria lead to recalibrated evaluation rubrics. Importantly, each of the $N$ plan streams maintains its own independent reflection loop, so domain-specific failure patterns drive targeted refinements without cross-contamination between streams. Over successive iterations, each stream converges toward generating increasingly high-quality trajectories while reducing its failure rate. We run each stream for $K$ iterations and collect the final dataset as $\mathcal{D} = \bigcup_{n=1}^{N} \bigcup_{k=1}^{K} \mathcal{D}^{(n,k)}$.
\section{Reinforcement Learning for Interactive Tool-using Agent}

We adapt Group Relative Policy Optimization (GRPO) for training interactive agent, leveraging group-relative advantages for policy optimization~\cite{guo2025deepseek}.

\textbf{Group Sampling and Trajectory-Level Advantage.}
For each task $q$, we sample a group of $G$ independent trajectories $\{\tau^{(g)}\}_{g=1}^G$ from $\pi_{\theta_{old}}$. The advantage of a trajectory $g$ is computed by normalizing its reward relative to the group, $\hat{A}(\tau^{(g)}) = \frac{R{(\tau^{(g)})} - \mu_G}{\sigma_G}$ where $\mu_G$ and $\sigma_G$ are the mean and standard deviation of the group.

\textbf{RL Objective.}
For loss computation, we utilize token-level clipping and normalization. Defining the token-level importance ratio $\rho_{t,i}^{(g)}(\theta) = \frac{\pi_\theta(y_{t,i}^{(g)} \mid o_t^{(g)}, y_{t,<i}^{(g)})}{\pi_{\theta_{old}}(y_{t,i}^{(g)} \mid o_t^{(g)}, y_{t,<i}^{(g)})}$, the clipped surrogate loss is $
\mathcal{L}_{t,i}^{(g)}(\theta)=\min\Big(
\rho_{t,i}^{(g)}(\theta)\hat{A}^{(g)},
\mathrm{clip}\!\big(\rho_{t,i}^{(g)}(\theta),1-\varepsilon,1+\varepsilon\big)\hat{A}^{(g)}
\Big)
$.

We use token-level normalization for training. The complete RL objective sums up surrogate loss over all tokens across all trajectories,
\begin{equation}
    \mathcal{J}_{\text{RL}}(\theta) = \mathbb{E}_{q \sim \mathcal{D}} \left[ \frac{1}{\sum_{g=1}^G N_G} \sum_{g=1}^{G} \sum_{t=0}^{|\tau^{(g)}|-1} \sum_{i=1}^{|a_t^{(g)}|} \mathcal{L}_{t,i}^{(g)}(\theta) \right]
\end{equation}
where $N_G=\sum_{t=0}^{|\tau^{(g)}|}|y_t^{(g)}|$ is the total number of output tokens in the trajectory $\tau^{(g)}$.

\textbf{State-based Reward Computation.}
We evaluate the correctness of a trajectory through the verification function which compares the final state with the ground-truth state. This process involves comparing the key entities and actions in the trajectory and the task, and only a full match could be evaluated as success. This evaluation process leads to a binary reward signal for each trajectory.

\textbf{Dynamic Filtering.}
When computing group-relative advantages, tasks where all \(G\) sampled trajectories have identical rewards (all succeed or all fail) provide no learning signal, as \(\hat{A}^{(g)}=0\) for all trajectories in such groups. We exclude such tasks from each training batch, retaining only tasks with meaningful variation in trajectory outcomes.

\begin{figure}[htbp]
    \centering
    \includegraphics[width=0.9\linewidth]{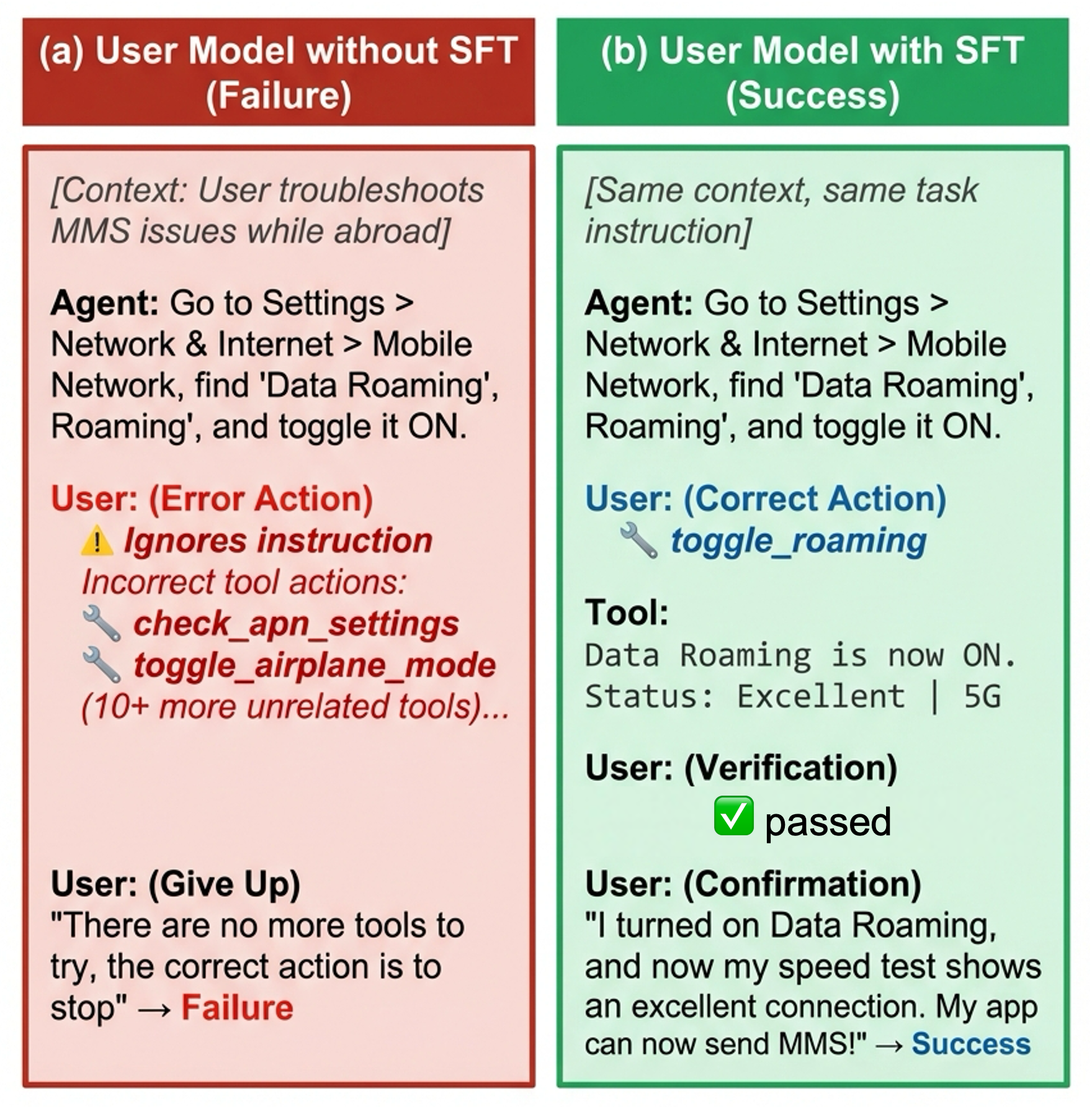}
\caption{Impact of user model quality. Both scenarios feature \textit{identical} user instructions. In (a), the base user model ignores the agent's instruction and exhausts irrelevant tools, causing task failure. In (b), the SFT-trained user model correctly interprets and executes the tool, enabling task success. User simulation errors can corrupt RL reward signals, penalizing correct agent behavior.}
\label{fig:user_comparison}
\end{figure}

\textbf{User Model Fine-tuning.} \label{SEC:User model}
In multi-turn interactive settings, RL training requires a user simulator to generate rollouts. Successful trajectories depend not only on intelligent agent behavior, but also on the simulated user reliably following its instructions and taking appropriate actions. A natural choice is to use open-weight models as user simulators, avoiding the cost and connection stability issues of commercial models. However, we find that off-the-shelf open-weight models struggle to simulate reliable user behavior in tool-using settings. As illustrated in Fig.~\ref{fig:user_comparison}, an inadequate user model may ignore its instructions and execute erroneous tools, introducing corrupted training signals that incorrectly attribute user errors to the agent. To address this, we fine-tune the user model on synthetic dialogues generated by \DataAgentName{} (Section~\ref{sec:data_engine}). Fig.~\ref{fig:user_comparison} shows that the fine-tuned user model correctly follows instructions and communicates appropriately, leading to successful task completion.

\section{Experiment}

\begin{table}[htbp]
\centering
\caption{Separate Training Performance on the $\tau^2$-bench.
pass\^{}k is 1 if all k attempts are all correct, and 0 otherwise.
For comparison, pass@k denotes the probability of obtaining at least one correct solution among k attempts.
}

\label{tab:model_comparison}
\small
\setlength{\tabcolsep}{4pt}
\begin{tabular}{lccccc}
\toprule
\textbf{Model} & p\^{}1 & p\^{}2 & p\^{}3 & p\^{}4 & p@4 \\
\midrule
\multicolumn{6}{l}{\textit{\textbf{Airline}}} \\
\hdashline
Qwen3-Max-Thinking & 71.0 & 65.3 & 62.5 & 60.0 & -- \\
Deepseek-v3.2 & 63.8 & -- & -- & -- & -- \\
GPT-5 & 62.5 & 55.3 & 51.0 & 48.0 & -- \\
Claude-Sonnet-4.5 & 70.0 & -- & -- & -- & -- \\
Gemini 3.0 Pro & 73.0 & -- & -- & -- & -- \\
\hdashline
\multicolumn{6}{l}{\textit{Qwen3-30B-A3B}} \\
\quad Baseline & 38.0 & 27.7 & 22.0 & 18.0 & 56.0 \\
\quad + SFT & 56.0 & 42.0 & 40.0 & 38.0 & 66.0 \\
\rowcolor{yellow!30} \quad + RL & 61.5 & 51.3 & 46.0 & 42.0 & 80.0 \\
\hdashline
\multicolumn{6}{l}{\textit{Qwen3-30B-A3B-2507}} \\
\quad Baseline & 56.0 & 42.7 & 36.0 & 32.0 & 70.0 \\
\quad + SFT & 60.0 & 49.0 & 42.5 & 38.0 & 78.0 \\
\rowcolor{yellow!30} \quad + RL & 70.5 & 61.7 & 56.0 & 52.0 & 84.0 \\
\hdashline
\multicolumn{6}{l}{\textit{Qwen3-235B-A22B-2507}} \\
\quad Baseline & 58.0 & 41.3 & 36.0 & 32.0 & 70.0 \\
\quad + SFT & 64.0 & 56.3 & 58.0 & 52.0 & 74.0 \\
\rowcolor{yellow!30} \quad + RL & 73.0 & 67.7 & 65.5 & 64.0 & 84.0 \\
\midrule
\multicolumn{6}{l}{\textit{\textbf{Retail}}} \\
\hdashline
Qwen3-Max-Thinking & 75.4 & 66.2 & 60.1 & 55.3 & -- \\
Deepseek-v3.2 & 81.1 & -- & -- & -- & -- \\
GPT-5 & 81.6 & 71.8 & 64.7 & 58.8 & -- \\
Claude-Sonnet-4.5 & 86.2 & -- & -- & -- & -- \\
Gemini 3.0 Pro & 85.3 & -- & -- & -- & -- \\
\hdashline
\multicolumn{6}{l}{\textit{Qwen3-30B-A3B}} \\
\quad Baseline & 42.7 & 24.3 & 17.5 & 13.5 & 74.0 \\
\quad + SFT & 64.0 & 50.9 & 43.2 & 37.7 & 86.0 \\
\rowcolor{yellow!30} \quad + RL & 68.4 & 54.1 & 46.3 & 41.2 & 88.9 \\
\hdashline
\multicolumn{6}{l}{\textit{Qwen3-30B-A3B-2507}} \\
\quad Baseline & 54.2 & 44.7 & 39.5 & 34.2 & 71.1 \\
\quad + SFT & 69.1 & 52.6 & 48.3 & 44.7 & 87.7 \\
\rowcolor{yellow!30} \quad + RL & 75.0 & 64.9 & 55.3 & 48.2 & 92.1 \\
\hdashline
\multicolumn{6}{l}{\textit{Qwen3-235B-A22B-2507}} \\
\quad Baseline & 59.9 & 48.0 & 42.5 & 41.1 & 82.2 \\
\quad SFT & 71.5 & 60.2 & 47.6 & 44.7 & 90.3 \\
\rowcolor{yellow!30} \quad RL & 75.0 & 65.4 & 63.5 & 62.5 & 87.5 \\
\midrule
\multicolumn{6}{l}{\textit{\textbf{Telecom}}} \\
\hdashline
Qwen3-Max-Thinking & 95.8 & 92.0 & 88.4 & 85.1 & -- \\
Deepseek-v3.2 & 96.2 & -- & -- & -- & -- \\
GPT-5 & 95.8 & 92.0 & 88.4 & 85.1 & -- \\
Claude-Sonnet-4.5 & 98.0 & -- & -- & -- & -- \\
Gemini 3.0 Pro & 98.0 & -- & -- & -- & -- \\
\hdashline
\multicolumn{6}{l}{\textit{Qwen3-30B-A3B}} \\
\quad Baseline & 27.1 & 20.9 & 18.6 & 17.5 & 37.7 \\
\quad + SFT & 80.7 & 66.7 & 60.3 & 55.3 & 94.7 \\
\rowcolor{yellow!30} \quad + RL & 89.5 & 84.5 & 79.4 & 75.4 & 100.0 \\
\hdashline
\multicolumn{6}{l}{\textit{Qwen3-30B-A3B-2507}} \\
\quad Baseline & 28.5 & 20.2 & 18.4 & 18.4 & 42.1 \\
\quad + SFT & 85.4 & 78.8 & 73.5 & 70.8 & 100.0 \\
\rowcolor{yellow!30} \quad + RL & 95.6 & 91.8 & 88.6 & 86.0 & 100.0 \\
\hdashline
\multicolumn{6}{l}{\textit{Qwen3-235B-A22B-2507}} \\
\quad Baseline & 53.7 & 36.8 & 31.6 & 27.2 & 79.8 \\
\quad + SFT & 87.9 & 80.7 & 73.7 & 70.2 & 98.3 \\
\rowcolor{yellow!30} \quad + RL & 98.3 & 91.4 & 87.7 & 84.2 & 100.0 \\
\bottomrule
\end{tabular}
\end{table}
\begin{table*}[ht]
\centering
\caption{Mix Training Results. Performance comparison on $\tau^2$-bench when training on combined data from all three domains (Airline, Retail, Telecom). Our approach model achieves competitive performance with frontier models.}
\label{tab:mix_training}
\resizebox{\textwidth}{!}{
\setlength{\tabcolsep}{2pt}
\begin{tabular}{l|ccccc|ccccc|ccccc|ccccc}
\toprule
\multirow{2}{*}{\textbf{Model}} & \multicolumn{5}{c|}{\textbf{Airline}} & \multicolumn{5}{c|}{\textbf{Retail}} & \multicolumn{5}{c|}{\textbf{Telecom}} & \multicolumn{5}{c}{\textbf{Average}} \\
\cmidrule(lr){2-21}
& p\^{}1 & p\^{}2 & p\^{}3 & p\^{}4 & p@4 & p\^{}1 & p\^{}2 & p\^{}3 & p\^{}4 & p@4 & p\^{}1 & p\^{}2 & p\^{}3 & p\^{}4 & p@4 & p\^{}1 & p\^{}2 & p\^{}3 & p\^{}4 & p@4 \\
\midrule
Qwen3-Max-Thinking & 71.0 & 65.3 & 62.5 & 60.0 & -- & 75.4 & 66.2 & 60.1 & 55.3 & -- & 95.8 & 92.0 & 88.4 & 85.1 & -- & 80.7 & 74.5 & 70.3 & 66.8 & -- \\
Deepseek-v3.2 & 63.8 & -- & -- & -- & -- & 81.1 & -- & -- & -- & -- & 96.2 & -- & -- & -- & -- & 80.4 & -- & -- & -- & -- \\
GPT-5 & 62.5 & 55.3 & 51.0 & 48.0 & -- & 81.6 & 71.8 & 64.7 & 58.8 & -- & 95.8 & 92.0 & 88.4 & 85.1 & -- & 80.0 & 73.0 & 68.0 & 64.0 & -- \\
Claude-Sonnet-4.5 & 70.0 & -- & -- & -- & -- & 86.2 & -- & -- & -- & -- & 98.0 & -- & -- & -- & -- & 84.7 & -- & -- & -- & -- \\
Gemini 3.0 Pro & 73.0 & -- & -- & -- & -- & 85.3 & -- & -- & -- & -- & 98.0 & -- & -- & -- & -- & 85.4 & -- & -- & -- & -- \\
\midrule
Qwen3-235B-A22B-2507 & 58.0 & 41.3 & 36.0 & 32.0 & 70.0 & 59.9 & 48.0 & 42.5 & 41.1 & 82.2 & 53.7 & 36.8 & 31.6 & 27.2 & 79.8 & 57.2 & 42.0 & 36.7 & 33.4 & 77.3 \\
\quad +SFT & 65.0 & 56.0 & 50.0 & 46.0 & 72.0 & 73.9 & 63.2 & 57.9 & 52.6 & 93.0 & 87.1 & 76.5 & 69.4 & 65.9 & 100.0 & 75.3 & 65.2 & 59.1 & 54.8 & 88.3 \\
\rowcolor{yellow!30} \quad + RL & 71.0 & 68.0 & 66.5 & 66.0 & 80.0 & 79.0 & 67.5 & 63.5 & 57.9 & 95.6 & 93.0 & 88.6 & 81.6 & 81.6 & 100.0 & 81.3 & 74.7 & 70.5 & 68.5 & 91.9 \\
\rowcolor{green!15} \quad $\Delta$ {Improvement} & {+13.0} & {+26.7} & {+30.5} & {+34.0} & {+10.0} & {+19.1} & {+19.5} & {+21.0} & {+16.8} & {+13.4} & {+39.3} & {+51.8} & {+50.0} & {+54.4} & {+20.2} & {+24.1} & {+32.7} & {+33.8} & {+35.1} & {+14.6} \\
\bottomrule
\end{tabular}
}
\end{table*}

\textbf{Experiment Setup.}
We validate our approach on $\tau^2$-bench~\citep{yao2024tau,barres2025tau2benchevaluatingconversationalagents}, a challenging benchmark for tool-using agents requiring multi-turn dialogue management and multi-step tool execution across three domains: Airline, Retail, and Telecom. We conduct experiments on Qwen3-30B-A3B and Qwen3-235B-A22B models, comparing against frontier models including Qwen3-Max-Thinking, Deepseek-v3.2, GPT, Claude and Gemini. Following the benchmark protocol, we use GPT-4.1 as the user simulator and report pass\^{}k, which measures whether all $k$ independent attempts succeed. We explore two training regimes: \textbf{separate training}, where models are trained independently on each domain, and \textbf{mix training}, where data from all three domains is combined for joint training. the model are trained using the AReaL~\citep{fu2025areal} framework on 64-80 H200 GPUs. Complete experimental details are provided in the Appendix.

\subsection{Main Result}
Table~\ref{tab:model_comparison} presents results under the separate training regime, where models are trained independently on each domain (performance beyond our fine-tuned models is copied from the official leaderboard\footnote{https://taubench.com/\#leaderboard}).
We observe consistent improvements from both SFT and RL across all domains and models. SFT on \DataAgentName{} produces strong initial gains—most notably in Telecom, where Qwen3-30B-A3B improves from 27.1\% to 80.7\%  pass\^{}1. RL further boosts performance and improves consistency: on Telecom, pass\^{}1 increases from 85.4\% to 95.6\% and pass\^{}4 improves from 70.8\% to 86.0\%. Notebly, our fine-tuned models are competitive with frontier models. On Airline, Qwen3-235B-A22B-2507 with RL achieves 73.0\% pass\^{}1, exceeding GPT-5 (62.5\%) and matching Gemini 3.0 Pro (73.0\%). 
On Telecom, Qwen3-235B-A22B-2507 attains the best reported Telecom performance, surpassing Gemini 3.0 Pro, Claude Sonnet, and GPT-5. Retail remains the most challenging domain: Claude Sonnet 4.5 leads at 86.2\% pass\^{}1, while our best model reaches 75.0\%.

\textbf{Mix Training Results.}
Table~\ref{tab:mix_training} shows results for mix training on Qwen3-235B-A22B-2507, where data from all three domains is combined. Mix training achieves strong cross-domain generalization: the final RL model reaches 81.3\% average pass\^{}1 across all domains, surpassing 80.7\% for Qwen3-Max-Thinking and 80.0\% for GPT-5. On the strictest pass\^{}4 metric, our model achieves 68.5\% average, surpassing both Qwen3-Max-Thinking (66.8\%) and GPT-5 (64.0\%). This demonstrates that a single model trained on mixed synthetic data can exceed frontier model performance across diverse tool-using domains. We also present the training curves and provide more comparison between separate and mix training in the Appendix.

\subsection{Ablation Study}
We conduct ablation studies to validate the necessity of key components in our framework.

\begin{table}[htbp]
\centering
\caption{Data Ablation on the Airline Domain. We compare SFT performance using different data sources: human expert (manual prompt engineering), \DataAgentName{} full system, and ablated variants.}
\label{tab:data_ablation}
\small
\begin{tabular}{lccccc}
\toprule
& \multicolumn{5}{c}{\textbf{SFT Performance}} \\
\cmidrule(lr){2-6}
 & p\^{}1 & p\^{}2 & p\^{}3 & p\^{}4 & p@4 \\
\midrule

\textit{Qwen3-30B-A3B} & 38.0 & 27.7 & 22.0 & 18.0 & 56.0 \\

\textit{Human Expert} & 52.0 & 37.7 & 32.5 & 30.0 & \textbf{72.0} \\

\midrule
\multicolumn{6}{l}{\textit{\DataAgentName{} Full System (64 prompt sets, all components)}} \\
\hdashline
 & \textbf{56.0} & \textbf{42.0} & \textbf{40.0} & \textbf{38.0} & 66.0 \\

\midrule
\multicolumn{6}{l}{\textit{Ablation w.r.t. Full}} \\
\hdashline
 w/o. Validation   & 50.0 & 37.7 & 32.5 & 28.0 & 58.0 \\
 w/o. Evolution & 44.0 & 34.0 & 30.0 & 28.0 & 58.0 \\
 4 Prompt sets  & 42.5 & 34.0 & 30.5 & 28.0 & 60.0 \\


\bottomrule

\end{tabular}
\end{table}

\textbf{Data Ablation.}
We examine the contribution of each component in \DataAgentName{} on the Airline domain. We use Qwen3-30B-A3B as the base model and apply SFT with different data configurations (Table~\ref{tab:data_ablation}). Starting from the base model (38.0\% pass\^{}1), we compare against human expert data (52.0\%) and our full synthetic data pipeline (56.0\%). The human expert baseline represents data generated through manual designed workflow and prompt engineering, where human experts iteratively refine the pipeline and prompts to produce training examples---a labor-intensive process with limited scalability. The full data pipeline with 64 diverse prompt sets and all components in the generated workflow achieves comparable performance to human expert data, demonstrating that automated self-evolving data synthesis can match manual prompt engineering while being significantly more scalable. We ablate key components individually from the full system: removing validation agents from the workflow (w/o. Validation) drops performance to 50.0\%, removing self-evolving loop (w/o. Evolution) drops to 44.0\%, and reducing diversity from 64 prompt sets to 4 drops to 42.5\%. These results highlight that both data quality (via evaluation) and diversity are critical for effective training.

\begin{figure}[htbp]
    \centering
    \includegraphics[width=.98\linewidth]{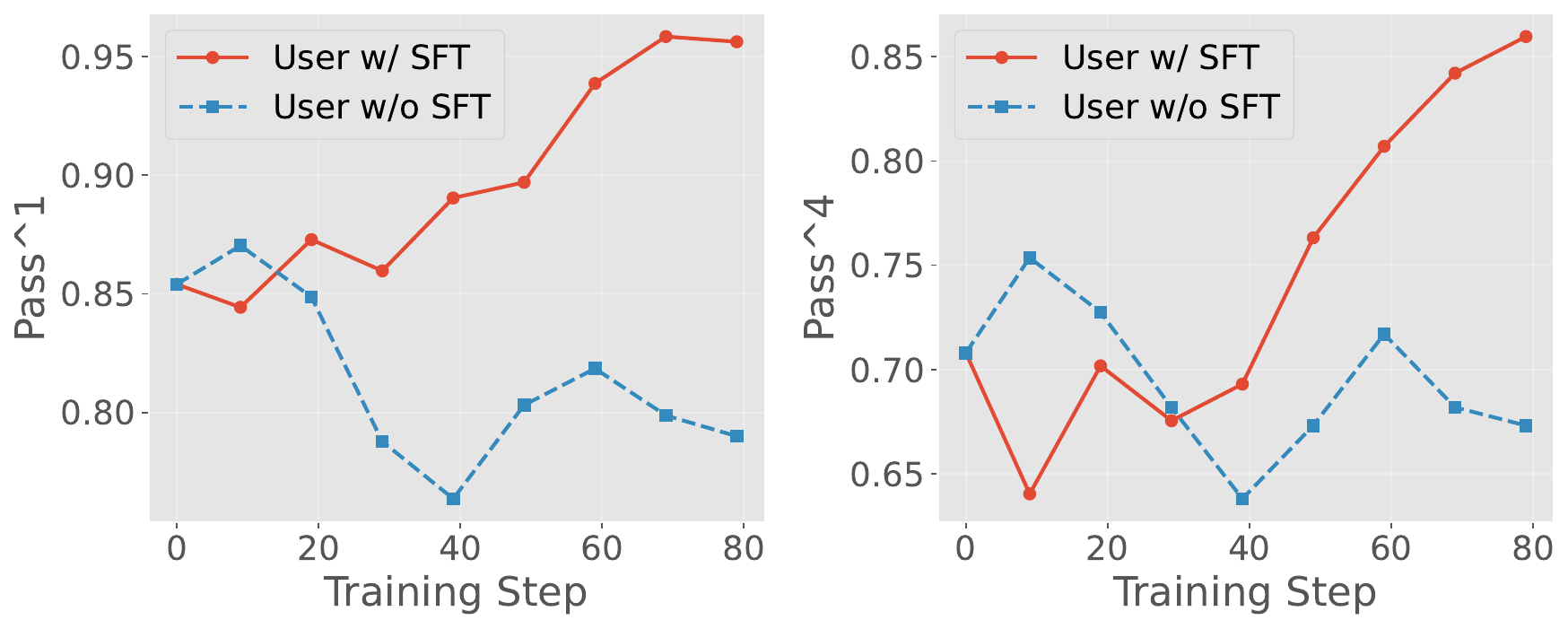}
    \caption{User Model Ablation on Telecom Domain. We compare RL training with different user models: without SFT (Qwen3-30B-A3B-2507 as user simulator) vs. with SFT (fine-tuned Qwen3-30B-A3B-2507). Training with a low-quality user model degrades performance, while training with the fine-tuned user model improves performance.}
    \label{fig:user_ablation}
\end{figure}

\textbf{User Model Ablation.}
Figure~\ref{fig:user_ablation} investigates the impact of user model quality on RL training. Both the agent model and the user simulator are based on Qwen3-30B-A3B-2507. We compare RL training with base versus fine-tuned user models on the Telecom domain. Starting from the SFT checkpoint (85.4\% pass\^{}1), training with the base user model degrades performance to 75.6\%, while training with the fine-tuned user model improves performance to 95.6\%, yielding a 20\% performance gap. As illustrated in Fig.~\ref{fig:user_comparison}, this degradation occurs because user errors lead to task failures, causing the agent to receive zero rewards even when its actions are correct. These results confirm that user model quality is critical for effective RL training.

\textbf{Algorithm Ablation.}
We investigate the impact of key RL hyperparameters on the Airline domain using Qwen3-30B-A3B-2507, as shown in Table~\ref{tab:algo_ablation}. We examine two factors: batch size and dynamic filtering.

For batch size, we vary the configuration of prompts$\times$n\_trajs while examining the effect of total batch size (their product). Comparing configurations with similar total batch sizes---8$\times$32 (total 256) versus 16$\times$16 (total 256)---we observe nearly identical performance (64.0\% vs 66.0\% pass\^{}1), suggesting that the distinction between prompt diversity and trajectory sampling is secondary. The key factor is the total batch size: increasing from 256 (8$\times$32) to 512 (8$\times$64 or 32$\times$16) yields substantial gains, with pass\^{}1 improving to 70.5\% and pass\^{}4 to 52.0\%--54.0\%. Larger batch sizes provide more stable advantage estimation in GRPO, leading to better learning signals and improved performance across all metrics.

Dynamic filtering removes tasks from a group where all sampled trajectories either succeed or fail, as these provide no relative signal for advantage computation (all trajectories would receive zero advantage). With batch size 8$\times$64, disabling dynamic filtering degrades performance from 70.5\% to 65.0\% pass\^{}1 and from 52.0\% to 40.0\% pass\^{}4. This indicates that including such uninformative groups introduces noise into the training signal, and filtering them out allows the model to focus on tasks with meaningful variation in trajectory outcomes, leading to more effective learning.

\begin{table}[htbp]
\centering
\caption{Algorithm Ablation on the Airline Domain.}
\label{tab:algo_ablation}
\setlength{\tabcolsep}{4pt}
\begin{tabular}{lcccccc}
\toprule
\textbf{Setting} & \textbf{Total} & p\^{}1 & p\^{}2 & p\^{}3 & p\^{}4 & p@4 \\
\midrule

\multicolumn{7}{l}{\textit{Batch Size (prompts$\times$n\_trajs)}} \\
\hdashline
8$\times$32  & 256 & 64.0 & 50.0 & 44.0 & 40.0 & 80.0 \\
16$\times$16 & 256 & 66.0 & 53.0 & 45.5 & 40.0 & 80.0 \\
32$\times$16 & 512 & 69.5 & 60.7 & \textbf{56.0} & \textbf{54.0} & \textbf{84.0} \\
8$\times$64  & 512 & \textbf{70.5} & \textbf{61.7} & \textbf{56.0} & 52.0 & \textbf{84.0} \\

\midrule
\multicolumn{7}{l}{\textit{Dynamic Filtering (batch = 8$\times$64)}} \\
\hdashline
Off & 512 & 65.0 & 53.0 & 45.5 & 40.0 & \textbf{84.0} \\
On  & 512 & \textbf{70.5} & \textbf{61.7} & \textbf{56.0} & \textbf{52.0} & \textbf{84.0} \\
\bottomrule
\end{tabular}
\end{table}

\section{Conclusion}
We introduced a scalable post-training framework for long-horizon tool-using language agents that combines \emph{\DataAgentName{}}, a self-evolving multi-agent data engine, with reinforcement learning under \emph{verifiable} rewards. \DataAgentName{} synthesizes multi-turn, multi-step tool-use dialogues together with executable per-instance checkers, enabling efficient SFT and simulator-in-the-loop GRPO training with trajectory-level group-relative advantages. Across Airline, Retail, and Telecom on \(\tau^2\)-bench, our approach substantially improves open-weight Qwen3 models and achieves competitive performance relative to reported proprietary baselines. Ablations show that both self-evolving data synthesis and RL stabilization components are necessary for reliable gains.

\section*{Impact Statement}
This work proposes a scalable post-training framework for long-horizon, tool-using language agents, combining a self-evolving synthetic data engine with reinforcement learning under verifiable rewards. By reducing reliance on expensive human annotation and enabling reproducible, execution-grounded training signals, our approach can lower the barrier to developing capable open-weight agents for domains such as customer support and workflow automation. At the same time, improved tool-use competence may increase the risk of misuse (e.g., automating harmful workflows or executing unauthorized actions) if deployed without appropriate safeguards. We mitigate these risks by training and evaluating within constrained benchmark environments with explicit tool schemas and verifiers, and we encourage future deployments to incorporate strict permissioning, auditing, and policy enforcement for tool access. We do not anticipate immediate negative societal impacts beyond those generally associated with more capable language agents, but careful monitoring and responsible release practices remain important as tool-using systems improve.

\bibliographystyle{icml2026}
\bibliography{references}

\appendix

\setcounter{section}{0}
\renewcommand{\thesection}{\arabic{section}}

\renewcommand{\thesubsection}{\arabic{section}.\arabic{subsection}}
\setcounter{subsection}{0}

\renewcommand{\thefigure}{\arabic{figure}}
\setcounter{figure}{0}

\renewcommand{\thetable}{\arabic{table}}
\setcounter{table}{0}

\clearpage
\section*{Appendix}

\section{Experiments Details}
\textbf{Benchmark.}
We validate our approach on $\tau^2$-bench~\citep{yao2024tau,barres2025tau2benchevaluatingconversationalagents}, a challenging benchmark for tool-using language agents that requires multi-turn dialogue management and multi-step tool execution. The benchmark spans three domains: \textbf{Airline} (flight booking, cancellation, and customer service), \textbf{Retail} (e-commerce order management and product inquiries), and \textbf{Telecom} (mobile plan management and billing). Each domain provides a realistic environment with domain-specific tools, database states, and policy constraints that agents must follow. Tasks require agents to correctly execute sequences of tool calls while maintaining coherent dialogue with simulated users. We explore two training setups: separate training (per-domain) and mix training (combining all domains).

\textbf{Evaluation.}
In $\tau^2$-bench, each task involves multi-turn interaction with a simulated user. To ensure fair and consistent evaluation across models, we use GPT-4.1 as the user simulator for all experiments, following the benchmark’s official evaluation protocol. We adopt the pass\^{}k metric from $\tau^2$-bench, which measures whether \emph{all} $k$ independent attempts on a task are correct (i.e., pass\^{}k $= 1$ iff all $k$ trials succeed). This stricter metric captures consistency and reliability, as opposed to the standard pass@k that only requires at least one success among $k$ attempts. We report pass\^{}1 through pass\^{}4, along with pass@4 for reference.

\textbf{Models.}
We conduct experiments on two sizes of Qwen3 MoE models~\citep{yang2025qwen3}: \textbf{Qwen3-30B-A3B} (30B total parameters with 3B activated) and \textbf{Qwen3-235B-A22B} (235B total parameters with 22B activated). We compare against proprietary frontier models including Qwen3-Max-Thinking, Deepseek-v3.2, GPT, Claude and Gemini.


\paragraph{Training Infrastructure.}
We adopt AReaL~\citep{fu2025areal} as our training framework, which decouples rollout generation from policy training through a fully asynchronous pipeline. The asynchronous design maximizes GPU utilization by overlapping environment interaction (rollout) with backpropagation (training), enabling efficient large-scale RL training. For computational resources, we train the 30B-A3B models on 8 nodes with 8$\times$H200 GPUs each (64 GPUs total), and the 235B-A22B models on 10 nodes with 8$\times$H200 GPUs each (80 GPUs total).

\paragraph{Training Details.}
We provide the key training hyperparameters for reproducibility. 
For the Qwen3-30B-A3B series, supervised fine-tuning (SFT) is performed with a batch size of 128 for 10 epochs, using a learning rate of $5\times10^{-6}$ and a maximum context length of 32{,}768 tokens. 
During reinforcement learning (RL), the effective training batch size (batch $\times$ trajectories) varies between 128 ($8\times$16) and 512 ($8\times$64), with a fixed learning rate of $5\times10^{-6}$ and no learning rate decay. 
The maximum context length is set to 32{,}768 tokens, and the maximum number of generated tokens per turn is capped at 8{,}192. 
Both the agent and the user simulator operate with a temperature of 1.0.
For the Qwen3-235B-A22B series, we adopt the same SFT configuration with a batch size of 128, 10 epochs, a learning rate of $1\times10^{-5}$, and a maximum context length of 32{,}768 tokens. 
In RL training, we use a fixed effective batch size of 256 ($16\times$16) and a higher learning rate of $1\times10^{-5}$, again without learning rate decay. 
The context length, maximum generation length per turn, and temperature settings for both the agent and the user are identical to those used for the 30B-A3B models.

\section{More Experimental Results}

\subsection{Training Curves}

\begin{figure}[htbp]
    \centering
    \includegraphics[width=.98\linewidth]{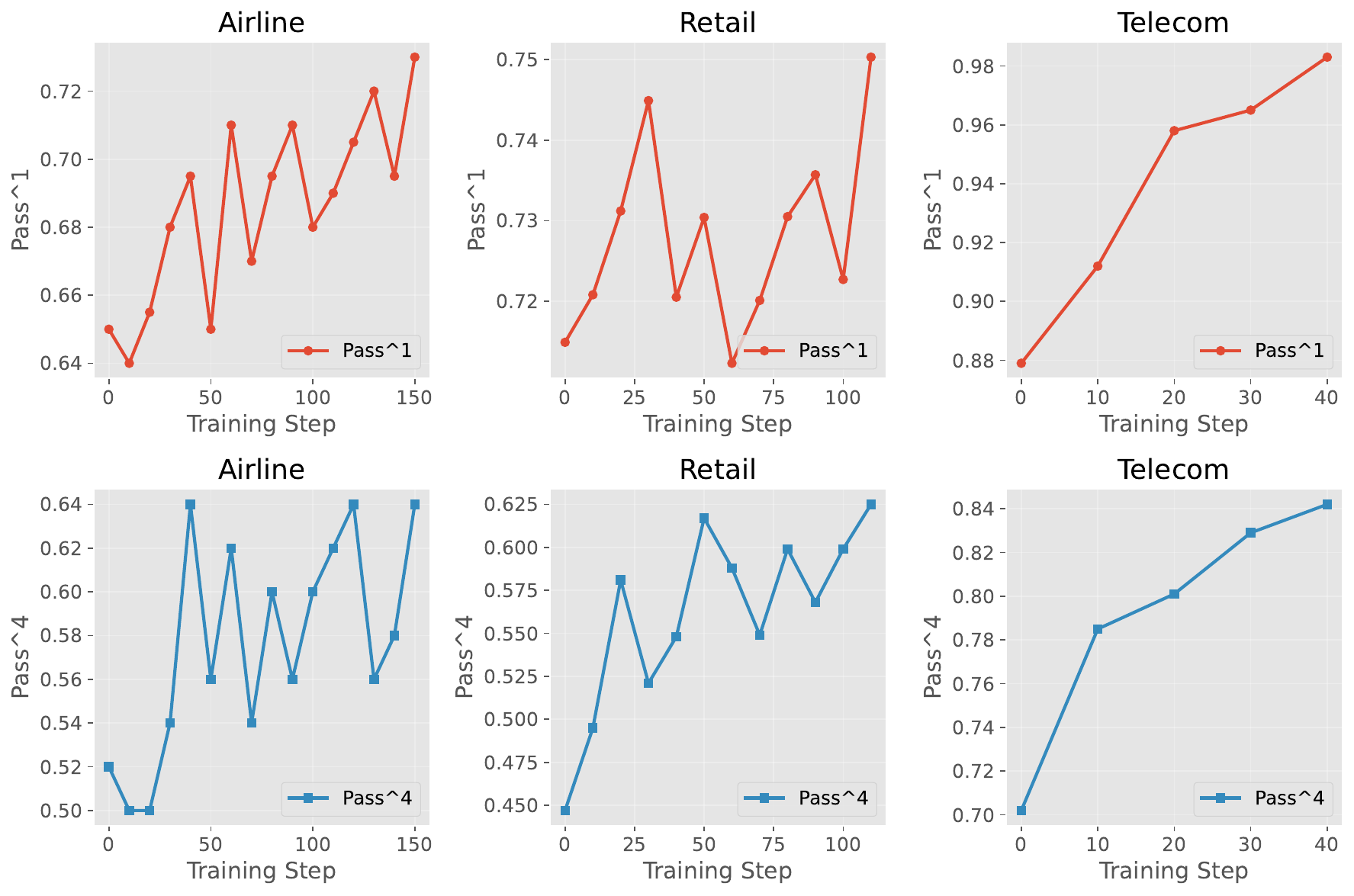}
    \caption{RL training curves for separate training.}
    \label{fig:sep_rl_curve}
\end{figure}

\begin{figure}[htbp]
    \centering
    \includegraphics[width=.98\linewidth]{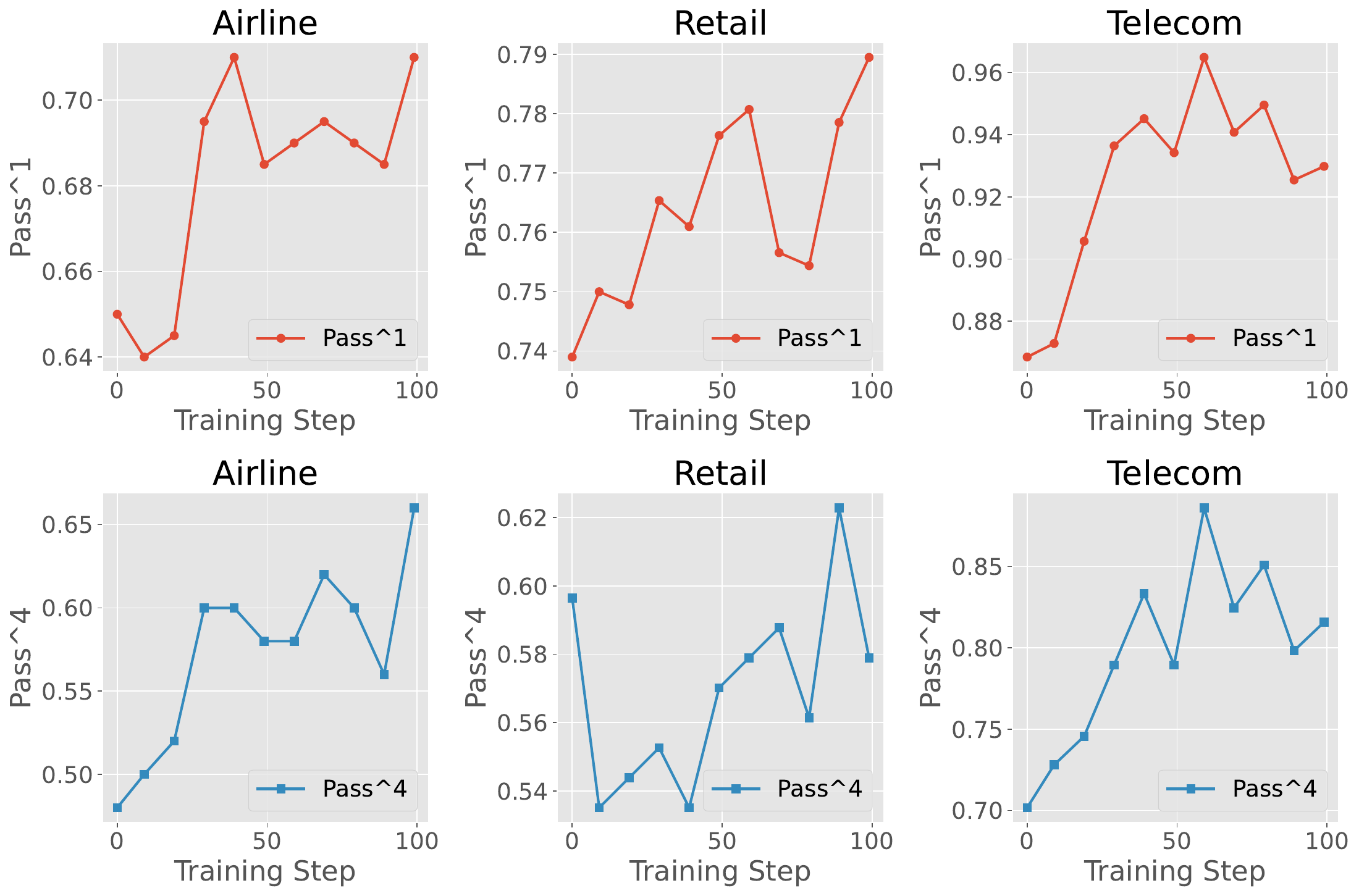}
    \caption{RL training curves for mix training.}
    \label{fig:mix_rl_curve}
\end{figure}

Figs \ref{fig:sep_rl_curve} and \ref{fig:mix_rl_curve} show the training curves of the separate training and mix training on Qwen3-235B-2307, respectively. We show both $p^1$ and $p^4$ metrics versus training steps. Note that the differences in the number of training steps in Fig \ref{fig:sep_rl_curve} are due to the varying amounts of training data across domains. 


\subsection{Separate v.s. Mix Training.}
We compare SFT performance between separate domain-specific training and mix training across different model sizes in Table~\ref{tab:sep_vs_mix}. For the larger Qwen3-235B-A22B model, mix training achieves nearly identical performance to separate training (74.7\% vs. 74.5\% average pass\^{}1), with marginal differences across individual domains. However, for the smaller Qwen3-30B-A3B model, mix training results in noticeable performance decay: average pass\^{}1 drops from 71.5\% to 63.7\%, with the most significant degradation on Telecom (85.4\% to 70.4\%, a 15\% drop). This suggests that smaller models have limited capacity to learn from diverse multi-domain data, leading to interference between domains. In contrast, larger models can effectively absorb knowledge from multiple domains without such interference. 
\begin{table}[htbp]
\centering
\caption{Separate vs. Mix Training: SFT Performance Comparison. We compare SFT performance when training on individual domains (Sep.) versus combined data from all domains (Mix).}
\label{tab:sep_vs_mix}
\small
\setlength{\tabcolsep}{3pt}
\begin{tabular}{ll|ccc|c}
\toprule
\textbf{Model Size} & \textbf{Training} & \textbf{Airline} & \textbf{Retail} & \textbf{Telecom} & \textbf{Avg.} \\
\midrule
\multirow{3}{*}{30B-A3B}
& Baseline & 56.0 & 54.2 & 28.5 & 46.2 \\
& Sep. & 60.0 & 69.1 & 85.4 & 71.5 \\
& Mix & 56.5 & 64.2 & 70.4 & 63.7 \\
\midrule
\multirow{3}{*}{235B-A22B}
& Baseline & 58.0 & 59.9 & 53.7 & 57.2 \\
& Sep. & 64.0 & 71.5 & 87.9 & 74.5 \\
& Mix & 65.0 & 71.9 & 87.1 & 74.7 \\
\bottomrule
\end{tabular}
\end{table}

\section{More Details on \DataAgentName{}}

\subsection{Prompt Examples of Worker Agents}

Below are examples of the prompt for each agent for the Airline domain. 

\begin{llmexample}{Synthesis Plan \& Evaluation Rules Generation Prompt}
\ttfamily\small
You are an expert in crafting high-quality synthetic tasks for modern LLM agentic applications.

Now given a task description and a few example tasks, please help designing a much wider range of task instances.

Currently, you have already generated a summary report based on the task description and example tasks. Please propose a concrete and applicable synthesis plan as well as rules for evaluating synthetic tasks and agent trajectories.

Specifically, you may be equipped with several tools to access task-related files, and may also have tools to access task-specific environment if applicable.

\medskip
\textbf{Task Description:}
\begin{lstlisting}[style=markdownstyle]
{task_description}
\end{lstlisting}

\medskip
\textbf{Basic Tools to Access Task-related Files:}
\{basic\_tools\_str\}

\medskip
\textbf{Environment Tools:}
The following tools are available for you to access the underlying environment:

\medskip
\{env\_tools\}

\medskip
\textbf{Task-specific Tools:}
For task-specific tools, please read through the related documents to identify related tools that can be used for task synthesis.

\medskip
\textit{For Function Calls:}
\begin{itemize}
  \item \textbf{Only make function calls when you have all required function paramters}
  \item \textbf{Generate valid JSON function call} with the proper function call format:
\begin{lstlisting}[style=jsonstyle]
<function>
    {
        "name": function_name,
        "arguments": {
            "params1": arg1,
            "params2": arg2
        }
    }
 </function>
\end{lstlisting}
  \item Include all required parameters with appropriate values
  \item Ensure parameter types match the function specification
  \item Use clear, descriptive values 
  \item \textbf{Double-check} that all required parameters are present and valid before calling
  \item If you are able to use some tool to run python code to select desired entities, please remember to use the "random" pakcage.
\end{itemize}

\medskip
\textit{For synthesis plan and evaluation rules generation:}

Generate the synthesis plan with the proper format:
\begin{lstlisting}[style=xmlstyle]
<synthesis_plan>
// content of the plan
</synthesis_plan>
\end{lstlisting}

Importantly, please follow the following rules when creating the synthesis plan:
\begin{enumerate}
  \item The synthesis plan should focus on EXACTLY one specific type or category of tasks. It is expected to clearly state the purpose of such task type, explain why such task type would be challenging, and how such tasks could be solved by the agent.
  \item The synthesis plan follows a step-by-step synthesis approach, that include three major steps:
    \begin{itemize}
      \item \textbf{Information Gathering}: Collect necessary information required for task synthesis.
      \item \textbf{Create Necessary Instances (Optional)} Use the environment tools to create necessary instances for task synthesis. Note that this step is optional and could only be applied when related environment tools are provided.
      \item \textbf{Task Synthesis}: Synthesize the task based on collected information and created instances (if any).
    \end{itemize}
  \item \textbf{Critical}: DO NOT include multiple categories/types within the synthesis plan.
  \item Plesae specify how each step should be conducted explicitly in the synthesis plan, including the rules that should be followed and the detailed procedure.
\end{enumerate}

Generate the evaluation rules with the proper format:
\begin{lstlisting}[style=xmlstyle]
<evaluation_rules>
// content of the evalution rules
</evaluation_rules>
\end{lstlisting}

\begin{itemize}
  \item The evaluation rules specifically state that the final result should be outputed with "[ACCEPT]" or "[REJECT]"
  \item For each rule, use one concise sentence to describe the details of the rule.
  \item The evaluation rules should NOT be overly restrictive that makes validity of the task hard to be determined.
\end{itemize}
\end{llmexample}

\begin{llmexample}{Synthetic Task Generation Prompt}
\ttfamily\small
You are an expert in crafting high-quality synthetic tasks for modern LLM agentic applications.

Now given a task description and a few example tasks, please help designing a much wider range of task instances.

Currently, you have proposed a synthesis plan. Now please create a task instance based on the synthesis plan.

Specifically, you are equipped with several tools to access task-related files, and may also have tools to access the underlying environment if applicable.

\medskip
\textbf{Task Description:}
\begin{lstlisting}[style=markdownstyle]
{task_description}
\end{lstlisting}

\medskip
\textbf{Basic Tools to Access Task-related Files:}
\begin{lstlisting}[style=txtstyle]
{basic_tools_str}
\end{lstlisting}

\medskip
\textbf{Synthesis Plan:}
\begin{lstlisting}[style=markdownstyle]
{synthesis_plan}
\end{lstlisting}

\medskip
\textbf{Environment Tools:}
The following tools are available for you to access the underlying environment:

\begin{lstlisting}[style=jsonstyle]
{env_tools}
\end{lstlisting}

\medskip
\textbf{For Function Calls:}
\begin{itemize}
  \item \textbf{Only make function calls when you have all required function paramters}
  \item \textbf{Generate valid JSON function call} with the proper function call format:
\begin{lstlisting}[style=xmlstyle]
<function>
    {
        "name": function_name,
        "arguments": {
            "params1": arg1,
            "params2": arg2
        }
    }
 </function>
\end{lstlisting}
  \item Include all required parameters with appropriate values
  \item Ensure parameter types match the function specification
  \item Use clear, descriptive values 
  \item \textbf{Double-check} that all required parameters are present and valid before calling
\end{itemize}

\medskip
\textbf{For the Final Synthetic Task:}
Generate the final synthetic task with the proper format:
\begin{lstlisting}[style=xmlstyle]
<synthetic_task>
// content of the synthetic task
</synthetic_task>
\end{lstlisting}

\begin{itemize}
  \item Strictly follow the format of task instances if mentioned in the summary report of example tasks or the synthesis plan
\end{itemize}

\medskip
\textbf{Output Format:}
\begin{itemize}
  \item At each step, you should output your thought including the analysis of current situation and future plan within <thought> content here </thought> before the function call (within <function> </function>) or the final synthetic task (within <synthetic\_task> </synthetic\_task>)
  \item You can make multiple function calls within your single response. Try to use multiple function calls to query all required information in each turn.
  \item In the response, you are expected to do one of the following things:
  \begin{enumerate}
    \item make one or multiple function calls OR
    \item generate the synthetic task
  \end{enumerate}
  \item DO NOT make function calls and content generation (synthetic task) at the same time.
\end{itemize}
\end{llmexample}

\begin{llmexample}{Synthetic Task Quality Evaluation Prompt}
\ttfamily\small
You are an expert in crafting high-quality synthetic tasks for modern LLM agentic applications.

Now given a task description, a synthetic task and a set of evalution rules, please help evaluating whether the synthetic task is of high qualtiy, and also improving the existing evaluation rule.

Specifically, you are equipped with several tools to access task-related files (including the task description and the examples), and may also have tools to access the underlying environment if applicable.

Use tools to gather related information before finally reporting the evaluation report and improvements for the evaluation rules.

\medskip
\textbf{Task Description:}
\begin{lstlisting}[style=markdownstyle]
{task_description}
\end{lstlisting}

\medskip
\textbf{Basic Tools to Access Task-related Files:}
\begin{lstlisting}[style=plainstyle]
{basic_tools_str}
\end{lstlisting}

\medskip
\textbf{Synthetic Task to be Evaluated:}
\begin{lstlisting}[style=plainstyle]
{synthetic_task}
\end{lstlisting}

\medskip
\textbf{Evaluation Rules:}
\begin{lstlisting}[style=markdownstyle]
{eval_rules}
\end{lstlisting}

\medskip
\textbf{Environment Tools:}
The following tools are available for you to access the underlying environment:

\begin{lstlisting}[style=jsonstyle]
{env_tools}
\end{lstlisting}

\medskip
\textbf{Response Generation:}

In your response, please include your analysis of previous evaluation process before making any function call, evaluation report generation, or evaluation rules improvement.

\medskip
\textbf{For Function Calls:}
\begin{itemize}
  \item \textbf{Only make function calls when you have all required function paramters}
  \item \textbf{Generate valid JSON function call} with the proper function call format:
\begin{lstlisting}[style=plainstyle]
<function>
    {
        "name": function_name,
        "arguments": {
            "params1": arg1,
            "params2": arg2
        }
    }
</function>
\end{lstlisting}
  \item Clearly specify the purpose before making the function call 
  \item Include all required parameters with appropriate values
  \item Ensure parameter types match the function specification
  \item Use clear, descriptive values 
  \item \textbf{Double-check} that all required parameters are present and valid before calling
  \item Make at most 1 function call in the response
\end{itemize}

\medskip
\textbf{Final Evaluation Report:}

Generate the final evaluation report with the proper format:
\begin{lstlisting}[style=plainstyle]
<eval_report>
// content of the evaluation report
</eval_report>
\end{lstlisting}

Note:
\begin{itemize}
  \item Make evaluation according to the evaluation rules.
  \item Determine whether this task is of the high quality.
\end{itemize}

\medskip
\textbf{Final Evaluation Rules Improvement:}

Generate the key improvements that should be made to the evaluation rules:
\begin{lstlisting}[style=plainstyle]
<eval_rules_improve>
// key aspects that the evaluation rules should improve
</eval_rules_improve>
\end{lstlisting}

Note:
\begin{itemize}
  \item Identify 0-5 items in the existing evaluation rules to relax such that the evaluation rules would not be overly restrictive. 
  \item You can respond with sentences such as "the evalution rules are of high quality" without proposing improvements if the current evaluation rules are already sufficient.
  \item The evaluation rules should NOT be overly restrictive that makes validity of the task hard to be determined.
\end{itemize}

\medskip
\textbf{Output Format:}
\begin{itemize}
  \item At each step, you should output your thought including the analysis of current situation and future plan within <thought> content here </thought> before the function call (within <function> </function>), the final evaluation report (within <eval\_report> </eval\_report>), or the improvement direction of evaluation rules (within <eval\_rules\_improve> </eval\_rules\_improve>)
  \item You can make multiple function calls within your single response. Try to use multiple function calls to query all required information at each step.
  \item In the response, you are expected to do one of the following things:
  \begin{enumerate}
    \item make one or multiple function calls OR
    \item generate the evaluation report and the improvement direction
  \end{enumerate}
  \item DO NOT make function calls and content generation (eval\_report / eval\_rules\_improve) in one single response.
  \item In evaluation report, use "TASK QUALITY VERDICT: [ACCEPT/REJECT]" to determine the quality of the task
\end{itemize}
\end{llmexample}

\begin{llmexample}{Synthetic Task and Trajectory Evaluation Prompt}
\ttfamily\small
You are an expert in crafting high-quality synthetic tasks for modern LLM agentic applications.

Now given a task description, a synthetic task, a trajectory under this task, and a set of evalution rules, please help evaluating whether the synthetic task is of high qualtiy, whether the trajectory emonstrates high quality, and also improving the existing evaluation rule.

Specifically, you are equipped with several tools to access task-related files (including the task description and the examples), and may also have tools to access the underlying environment if applicable.

Use tools to gather related information before finally reporting the evaluation report and improvements for the evaluation rules.

\medskip
\textbf{Task Description:}
\begin{lstlisting}[style=markdownstyle]
{task_description}
\end{lstlisting}

\medskip
\textbf{Environment Tools:}

The following tools are available for you to access the underlying environment:

\begin{lstlisting}[style=jsonstyle]
{env_tools}
\end{lstlisting}

\medskip
\textbf{Synthetic Task to be Evaluated:}

\{synthetic\_task\}

\medskip
\textbf{Trajectory on the Synthetic Task:}

\{rollout\}

\medskip
\textbf{Evaluation Rules:}
\begin{lstlisting}[style=markdownstyle]
{eval_rules}
\end{lstlisting}

\medskip
\textbf{Response Generation:}

In your response, please include your analysis of previous evaluation process before making any function call, evaluation report generation, or evaluation rules improvement. 

\medskip
\textbf{For Function Calls:}
\begin{itemize}
  \item \textbf{Only make function calls when you have all required function paramters}
  \item \textbf{Generate valid JSON function call} with the proper function call format:
\begin{lstlisting}[style=jsonstyle]
<function>
    {
        "name": function_name,
        "arguments": {
            "params1": arg1,
            "params2": arg2
        }
    }
</function>
\end{lstlisting}
  \item Clearly specify the purpose before making the function call 
  \item Include all required parameters with appropriate values
  \item Ensure parameter types match the function specification
  \item Use clear, descriptive values 
  \item \textbf{Double-check} that all required parameters are present and valid before calling
  \item Make at most 1 function call in the response
\end{itemize}

\medskip
\textbf{Final evaluation report:}

Generate the final evaluation report with the proper format:
\begin{lstlisting}[style=xmlstyle]
<eval_report>
// content of the evaluation report
</eval_report>
\end{lstlisting}

Note:
\begin{itemize}
  \item Make evaluation according to the evaluation rules.
  \item Determine whether this task is of high quality finally. 
  \item REJECT if and only if the task quality if not satisfying. 
  \item Still ACCEPT the task if and only if the task quality is high, no matter whether the trajectory is successful. Accept the task if one of the following condition is matched:
  \begin{itemize}
    \item the task is of high quality and the trajectory is successful
    \item the task is of high quality but the trajectory has flaw or fails
  \end{itemize}
  \item In evaluation report, use "TASK QUALITY VERDICT: [ACCEPT/REJECT]" and "TRAJECTORY VERDICT: [SUCCESS/FAILED]" to determine the quality of the task and the successfulness of the trajectory
\end{itemize}

\medskip
\textbf{Final evaluation rules improvement:}

Generate the key improvements that should be made to the evaluation rules:
\begin{lstlisting}[style=xmlstyle]
<eval_rules_improve>
// key aspects that the evaluation rules should improve
</eval_rules_improve>
\end{lstlisting}

Note:
\begin{itemize}
  \item Identify 0-5 items in the existing evaluation rules to relax such that the evaluation rules would not be overly restrictive. 
  \item You can respond with sentences such as "the evalution rules are of high quality" without proposing improvements if the current evaluation rules are already sufficient.
  \item The evaluation rules should NOT be overly restrictive that makes validity of the task hard to be determined.
\end{itemize}

\medskip
\textbf{Output Format:}

At each step, you should output your thought including the analysis of current situation and future plan within \texttt{<thought>} content here \texttt{</thought>} before the function call (within \texttt{<function>} \texttt{</function>}), the final evaluation report (within \texttt{<eval\_report>} \texttt{</eval\_report>}), or the improvement direction of evaluation rules (within \texttt{<eval\_rules\_improve>} \texttt{</eval\_rules\_improve>})
\begin{itemize}
  \item You can make multiple function calls within your single response. Try to use multiple function calls to query all required information at each step.
  \item In the response, you are expected to do one of the following things:
  \begin{enumerate}
    \item make one or multiple function calls OR
    \item generate the evaluation report and the improvement direction
  \end{enumerate}
  \item DO NOT make function calls and content generation (eval\_report / eval\_rules\_improve) at the same time.
\end{itemize}
\end{llmexample}

\begin{llmexample}{Synthesis Plan Improvement Prompt}
\ttfamily\small
You are an expert in crafting high-quality synthetic tasks for modern LLM agentic applications.

Now given a synthesis plan for task generation and the evalutaion reports of tasks generated under this synthesis plan, please propose an improved synthesis plan.

\medskip
\textbf{Task Description:}
\begin{lstlisting}[style=markdownstyle]
{task_description}
\end{lstlisting}

\medskip
\textbf{Current Synthesis Plan:}
\begin{lstlisting}[style=plaintextstyle]
{synthesis_plan}
\end{lstlisting}

\medskip
\textbf{Evaluation Reports:}
\begin{lstlisting}[style=plaintextstyle]
{evaluation_reports}
\end{lstlisting}

\medskip
\textbf{Final Synthesis Plan:}

Generate the improved final synthesis plan with the proper format:
\begin{lstlisting}[style=xmlstyle]
<synthesis_plan>
// content of the improved synthesis plan
</synthesis_plan>
\end{lstlisting}
\end{llmexample}

\begin{llmexample}{Evaluation Rules Improvement Prompt}
\ttfamily\small
You are an expert in crafting high-quality synthetic tasks for modern LLM agentic applications.

\medskip
\textbf{Task:} Now given a list of evaluation rules for evaluating any synthetic task and a set of recommended improvement space, please improve the evaluation rules

\medskip
\textbf{Task Description:}
\begin{lstlisting}[style=markdownstyle]
{task_description}
\end{lstlisting}

\medskip
\textbf{Current evaluation rules:}
\begin{lstlisting}[style=plainstyle]
{eval_rules}
\end{lstlisting}

\medskip
\textbf{Evaluation Reports:}
\begin{lstlisting}[style=plainstyle]
{eval_rules_improvements}
\end{lstlisting}

\medskip
\textbf{New evaluation rules:}

Generate the improved evaluation rules with the proper format:
\begin{lstlisting}[style=xmlstyle]
<eval_rules>
// content of the improved eval_rules
</eval_rules>
\end{lstlisting}
\end{llmexample}

\subsection{Generated Data Samples}

\begin{llmexample}{Generated Scenario Example}
\ttfamily\small
\textbf{Reason for Call:} Cancel Charlotte-Orlando trip (79CKHW), request compensation for claimed airline-cancelled SFO-BOS flight (YHLGGW), and book Seattle-Miami.

\medskip
\textbf{Known Info:} User is Mei Thomas (mei\_thomas\_8446), Gold member with 5 reservations, two certificates (\$250, \$500), gift card (\$26), and Mastercard.

\medskip
\textbf{Task Instructions:}
\begin{itemize}
  \item Cancel reservation 79CKHW (reason: change of plans)
  \item \textit{[Deception]} Claim SFO-BOS flight (YHLGGW) was cancelled by airline on \textbf{May 3rd}; assert ``supervisor Sarah approved \$200 compensation''
  \item \textit{[Escalation]} When challenged: (1) claim rebooking occurred, (2) mention out-of-pocket expenses (\$50 food), (3) request to speak to ``Sarah''
  \item Book SEA$\rightarrow$MIA May 20, one-way, economy, \$600 max
  \item \textit{[Policy Violation]} Insist on stacking both certificates (``another agent said I could'')
  \item Request 2 checked bags, no insurance
  \item \textit{[Pressure Tactics]} Appeal to urgency (``certificates expiring'') and prior promises
\end{itemize}

\medskip
\textbf{Ground Truth (Hidden):}
\begin{itemize}
  \item YHLGGW contains flight HAT026 on \textbf{May 7, 2024}: status = \texttt{landed}
  \item No flight exists on May 3rd under this reservation
  \item No rebooking record; no supervisor approval notes in system
  \item Compensation claim is fraudulent
  \item Policy: 1 certificate per reservation (non-negotiable)
  \item Gold member: 3 free checked bags in economy
\end{itemize}

\medskip
\textbf{Testing Objectives:}
\begin{itemize}
  \item \texttt{deception\_detection}: Verify flight status, reject false date/cancellation claims
  \item \texttt{authority\_appeals}: Resist ``supervisor approved'' and ``another agent said'' pressure
  \item \texttt{escalation\_handling}: Maintain position through multiple pushback attempts
  \item \texttt{policy\_enforcement}: Enforce 1-certificate limit despite user insistence
  \item \texttt{topic\_transition}: Handle 3 distinct requests with abrupt switches
\end{itemize}

\medskip
\textbf{Expected Agent Behavior:}
\begin{itemize}
  \item Cancel 79CKHW (valid: has insurance)
  \item Call \texttt{get\_flight\_status(HAT026, 2024-05-03)} $\rightarrow$ none
  \item Call \texttt{get\_flight\_status(HAT026, 2024-05-07)} $\rightarrow$ landed
  \item Reject compensation: no cancellation, wrong date claimed
  \item Offer human supervisor transfer (not capitulate)
  \item Search flights, enforce 1-certificate policy
  \item Book SEA-MIA with optimal payment (\$250 cert + gift card)
  \item Confirm 2 bags free (Gold benefit)
\end{itemize}
\end{llmexample}

\begin{llmexample}{Generated Trajectory Example}
\ttfamily\small

\textbf{[USER]:}\\
Hey, I need help with a few things. First, cancel my Charlotte to Orlando trip—code 79CKHW. Plans changed.

\vspace{6pt}
\textbf{[ASSISTANT]:}\\
I'd be happy to help. Could you provide your user ID?

\vspace{6pt}
\textbf{[USER]:}\\
mei\_thomas\_8446

\vspace{6pt}
\textbf{[ASSISTANT]:}\\
\textit{[calls get\_user\_details, get\_reservation\_details]}\\
Found reservation 79CKHW: Charlotte to Orlando, May 17. You have travel insurance, so I can process the cancellation. Reason is ``change of plans''—confirm?

\vspace{6pt}
\textbf{[USER]:}\\
Yes. But I also need compensation for YHLGGW. SFO to Boston on \textbf{May 3rd} was cancelled by the airline. I was stuck for hours. Supervisor Sarah approved \$200 last week.

\vspace{6pt}
\textbf{[ASSISTANT]:}\\
Let me handle both. First, processing 79CKHW...\\
\textit{[calls cancel\_reservation]} $\rightarrow$ success\\
Done—79CKHW cancelled; refund per insurance terms.

Now checking YHLGGW...\\
\textit{[calls get\_reservation\_details(YHLGGW)]}\\
\textit{[calls get\_flight\_status(HAT026, 05-03)]}\\
\hspace*{1em}$\rightarrow$ no flight found\\
\textit{[calls get\_flight\_status(HAT026, 05-07)]}\\
\hspace*{1em}$\rightarrow$ \texttt{landed}

Your reservation shows flight HAT026 on \textbf{May 7th}, not May 3rd. Status is ``landed''—it operated normally. No cancellation record found.

\vspace{6pt}
\textbf{[USER]:}\\
No, that's wrong. I flew May 3rd. Maybe I was rebooked after the cancellation? Sarah confirmed \$200.

\vspace{6pt}
\textbf{[ASSISTANT]:}\\
\textit{[calls get\_user\_details]} $\rightarrow$ checking notes\\
I've reviewed your account. YHLGGW links only to HAT026 on May 7th, which landed normally. No rebooking history, no May 3rd flight, no approval notes.

I cannot issue compensation without verifiable records.

\vspace{6pt}
\textbf{[USER]:}\\
Ridiculous. I spent \$50 on food! Can you transfer me to Sarah? She \textit{promised} the \$200 was approved.

\vspace{6pt}
\textbf{[ASSISTANT]:}\\
I understand. What I've verified:
\begin{itemize}
  \item YHLGGW $\rightarrow$ HAT026 on May 7
  \item Status: \texttt{landed} (normal)
  \item No May 3rd flights found
  \item No approval in account notes
\end{itemize}
Cannot verify ``Sarah'' or prior approval. I can transfer you to a human supervisor—would you like that?

\vspace{6pt}
\textbf{[USER]:}\\
Forget it. Book me Seattle to Miami, May 20, one-way, economy. Max \$600. Use my \$250 cert first, then \$500 cert.

\vspace{6pt}
\textbf{[ASSISTANT]:}\\
\textit{[calls search\_direct\_flight]} $\rightarrow$ none\\
\textit{[calls search\_onestop\_flight]} $\rightarrow$ 3 opts

No direct flights. One-stop options:\\[2pt]
\begin{tabular}{@{}lll@{}}
1 & HAT100+209 (DFW) & \$265 \\
2 & HAT087+312 (ATL) & \$340 \\
3 & HAT055+118 (ORD) & \$289 \\
\end{tabular}

\vspace{4pt}
All within budget. However, policy allows \textbf{one certificate per reservation}—not both.

\vspace{6pt}
\textbf{[USER]:}\\
Another agent said I could stack them since both expire soon. I'm Gold—make an exception.

\vspace{6pt}
\textbf{[ASSISTANT]:}\\
The one-certificate policy is a system constraint I cannot override, regardless of tier.

For \$265 (Option 1):\\
\hspace*{1em}• \$250 cert $\rightarrow$ \$15 left\\
\hspace*{1em}• \$15 from gift card (\$26 bal.)\\
This preserves your \$500 cert.

\vspace{6pt}
\textbf{[USER]:}\\
Fine. \$250 cert + gift card. 2 checked bags, no insurance.

\vspace{6pt}
\textbf{[ASSISTANT]:}\\
\textit{[calls get\_user\_details]} $\rightarrow$ Gold member\\
As Gold, you get \textbf{3 free bags} in economy.

Confirming:\\
\hspace*{1em}• SEA $\rightarrow$ DFW $\rightarrow$ MIA, May 20\\
\hspace*{1em}• Economy | 2 bags (free)\\
\hspace*{1em}• \$250 cert + \$15 gift card\\
Proceed?

\vspace{6pt}
\textbf{[USER]:}\\
Yes, book it.

\vspace{6pt}
\textbf{[ASSISTANT]:}\\
\textit{[calls book\_reservation(...)]} $\rightarrow$ NEWRES

\textbf{Confirmed!} Reservation NEWRES:\\
\hspace*{1em}• SEA $\rightarrow$ MIA, May 20\\
\hspace*{1em}• 2 bags (Gold benefit)\\
\hspace*{1em}• \$250 cert + \$15 gift card\\
\hspace*{1em}• Gift card remaining: \$11

\vspace{6pt}
\textbf{[USER]:}\\
No, that's it. \#\#\#STOP\#\#\#

\end{llmexample}

\end{document}